%% file: main.tex
\documentclass[sigconf,screen]{acmart}

\makeatletter                   
\def\mdseries@tt{m}             
\makeatother

\AtBeginDocument{%
  \providecommand\BibTeX{{%
    \normalfont B\kern-0.5em{\scshape i\kern-0.25em b}\kern-0.8em\TeX}}}

\usepackage{caption}
\usepackage{subcaption}
\usepackage{tabularx}
\usepackage{tikz-cd}
\usepackage{tikz}
\usepackage[draft=true]{minted} 
\usepackage{csquotes}
\usepackage{algorithm}
\usepackage{algpseudocode}
\usepackage{hyperref}           

\usetikzlibrary{arrows}

\begin{document}
\title{Why should I {\em not} follow you? 
Reasons For and Reasons Against in Responsible Recommender Systems}

\author{Gustavo P. Polleti}
\affiliation{\institution{Universidade de S\~ao Paulo}}
\email{gustavo.polleti@usp.br}

\author{Douglas L. de Souza}
\orcid{https://orcid.org/0000-0003-1019-2299}
\affiliation{\institution{Universidade de S\~ao Paulo}}
\email{douglas.luan.souza@usp.br}

\author{Fabio G. Cozman}
\affiliation{\institution{Universidade de S\~ao Paulo}}
\email{fgcozman@usp.br}

\acmConference[FAccTRec2020]{3rd FAccTRec Workshop: Responsible Recommendation}{September 26, 2020}{Online}
\acmBooktitle{RecSys '20: Proceedings of the 14th ACM Conference on Recommender Systems}
\setcopyright{none}
\copyrightyear{2020}
\acmYear{2020}

\begin{abstract}
A few Recommender Systems (RS) resort to explanations so as to enhance 
trust in recommendations. However, current 
techniques for explanation generation tend to strongly uphold the
recommended products instead of presenting both reasons for and
reasons against them. We argue that an RS can better enhance  
overall trust and transparency by frankly displaying both kinds of reasons 
to users. We have developed such an RS by exploiting knowledge graphs and
by applying Snedegar's theory of practical reasoning.
We show that our implemented RS has excellent performance and we 
report on an experiment with human subjects that shows the 
value of presenting both reasons for and against, with significant
improvements in trust, engagement, and persuasion.
\end{abstract}

\maketitle

\input{sections/1intro}

\input{sections/2background}

\input{sections/3proposal}

\input{sections/4experiments}
\input{sections/5conclusion}
\input{sections/6acknowledgements}

\bibliographystyle{ACM-Reference-Format}
\bibliography{main}

\end{document}

%% file: sections/1intro.tex
\section{Introduction}
\label{sec:intro}

Human subjects find it hard to make a decision when a very 
large number of options is available; a Recommender System (RS) provides valuable help by selecting a small set of options that 
are then evaluated by the user~\cite{Ricci2015}. However, even if an RS 
produces sensible recommendations, users may reject them if 
their rationale is not understood~\cite{Sinha:2002:RTR:506443.506619}.
It is thus clearly desirable to have RSs that offer sensible, 
transparent and trustworthy recommendations; one strategy that seems
particularly promising is for the RS to generate explanations that 
clarify the recommendations~\cite{Tintarev:2007:SER:1547550.1547664}.

Explanations presumably enhance transparency and trust. However, 
explanation generation techniques  now in use in RSs focus solely on 
advocacy for the recommended options. By describing only the
benefits of those options, they may fail to offer a balanced perspective
to the user, ultimately squandering overall trust. A user may be
at first happy to get some positive clarification about recommended products,
but if she never sees  information about possible downsides, she will ultimately lose interest in the recommendations. 

We argue that an RS should provide {\em responsible}
explanations in the sense that both reasons {\em for} and 
reasons {\em  against} explicitly escort recommendations. 
We take Snedegar's theory of reasons for/against \cite{snedegar2018reasons},
a philosophical theory of practical reasoning, and realize it in the
context of RSs. To do so, we start with existing procedures that
generate reasons {\em for} by analyzing paths in knowledge graphs
\cite{SemanticsMARecsys19,MUSTO201993,POLLETIWICRS20}.
We then modify such procedures so as to detect paths (or their absence) 
that count as reasons {\em against}.
Snedegar's theory relies on five schemes of reasons against;
we examine their computational implementation, identifying the most
promising strategies. We also describe an RS we have implemented 
and its practical operation with reasons for/against. 
Additionally, we have carried out experiments
with human subjects that show our approach to responsible recommendations to yield 
higher overall trust in generated explanations.

The paper is organized as follows. Section \ref{sec:back} presents
some basic notions on recommender systems, explainability, transparency, and trust. 
In Section \ref{sec:proposal} we propose strategies to generate reasons
for/against. 
We then present our empirical results, and offer concluding
remarks in the last section.

%% file: sections/2background.tex
\section{A bit of background: Recommendations, Trust, Interpretability, Explanations}
\label{sec:back}

An RS has a set of users and a set
of items, usually producing a score $r(u,i)$ that captures
the affinity between user $u$ and item $i$~\cite{Ricci2015}.
An RS often relies on the score to rank a number $N$ of items 
to be presented to the user.
The definition of affinity varies wildly,
depending on the application domain~\cite{MF09,10.5555/2981562.2981720}. 
The current state of the art is to learn the affinity between users and
items from past experience using latent variable models, often
dependent on matrix factorization and embedding techniques~\cite{He2017,Henk2018,Huang2019}. 
These techniques map items to a (numeric) latent space where similar 
items appear near to each other, usually by optimizing distances
between related objects as they are mapped~\cite{Nickel2015}. 

Opaque models, such as the ones produced by embeddings,
create obstacles to the interpretability of 
recommendations~\cite{doshivelez2017towards}. Here we take interpretability
as the degree to which a human can understand the cause of a decision \cite{Miller2017}. 
A device may be transparent in that the user can access
all elements of its operation, yet its output may have low interpretability.
When interpretability is low, one possible strategy is to generate
explanations for the decisions.
There are several techniques for explanation generation~\cite{molnar2019}; 
for instance, some of them investigate the sensitivity of outputs to  
inputs or to elements of a model --- the explanation is an indication
of which parts of input/model affect the output. Other techniques aim at
more elaborate explanations. Some of them are dependent on a particular model;
for instance, some techniques focus on neural networks, producing explanations
that involve particular neurons and layers. Other techniques for explanation
generation are model agnostic; that is,
they only look at inputs and outputs of the model to be explained.
We focus on model-agnostic explanations in this work.

It is commonly stated that performance and interpretability 
are opposing goals \cite{ribeiro16}; for instance, an accurate classifier is 
a complex and hard to interpret one.
However, matters are more delicate in the context of RSs, as 
performance itself depends on trust~\cite{Donovan2005}, 
and high interpretability is bound to increase trust (when
interpretation fails, existing RSs may fail in surprising ways \cite{whenRecSysGoBad16}).
Previous efforts have explored various ways to obtain high
performance and high interpretability~\cite{10.1145/1297231.1297235,10.1145/1935826.1935877,DBLP:journals/corr/abs-1804-11192}, in some cases generating explanations 
that support recommendations~\cite{MUSTO201993, SemanticsMARecsys19, LISTEN18}.

%% file: sections/3proposal.tex
\section{Explanations with  Reasons For and Reasons Against}
\label{sec:proposal}

Recent RSs that rely on explanations do offer useful information
to the user; however, we argue that they run into a difficult
balancing act~\cite{Milano2020}. This is not unlike the salesperson
who always proposes products with complimentary words, as opposed
to the salesperson who frankly discusses the advantages and disadvantages
of products. 
A perceptive customer will gradually favor a salesperson who
chooses sincerity over persuasion --- exactly the  behavior
we propose for responsible RSs.

The solution, then, is to build RSs that state reasons {\em for}
recommended items together with reasons {\em against} the same items.
This is the main idea in this paper; to make it concrete, we 
first discuss techniques that generate reasons {\em for} 
(Section \ref{sec:reasons_for}) and then we propose novel ideas
on the generation of reasons {\em against} (Section~\ref{sec:reasons_against}).

\input{sections/31reasonsfor}

\input{sections/32reasonsagainst}

%% file: sections/31reasonsfor.tex
\subsection{Reasons For: What They Are, and How to Generate Them}
\label{sec:reasons_for}

Reasons for a given recommendation can be produced using an auxiliary knowledge graph (KG), 
a strategy that has been 
explored in previous efforts~\cite{MUSTO201993,SemanticsMARecsys19,POLLETIWICRS20}.

The idea is to use a KG containing all entities handled by the RS so as to 
find connections between users and items.
A knowledge graph (KG) consists of a set of {\em entities} $\mathcal{E} = \lbrace e_{1}, \dots, e_{N_{e}} \rbrace $  and a set of binary {\em relations} $\mathcal{R} = \lbrace r_{1}, \dots, r_{N_{r}} \rbrace $. Using  RDF  notation \cite{RDFframe}, an edge in the graph can be interpreted as a triple $\langle h , r, t \rangle $ where $h$, $r$ and $t$ are, respectively, the \textit{subject (head)}, \textit{predicate (relation)} and \textit{object (tail)}. The existence of a triple $x_{h, r, t} = \langle h , r, t \rangle$ is indicated by a random variable $y_{h,r,t}$ with values in $\lbrace 0, 1 \rbrace$.
A {\em path type} $\pi$ is a sequence of relations
$r_1-r_2-...-r_l$, some of which may be the inverses of relations in $\mathcal{R}$
(the inverse of relation $r$ is denoted by $r^-$).
A given path $\pi$ {\em holds} for entities $h$ and $t$ if there exists a set of entities ${e_{1}, e_{2}, ...}$ so that all the variables $\{ y_{h,r_{1},e_{1}}, y_{e_{1},r_{2},e_{2}}, ... y_{e_{l-1},r_{l},t} \}$ have value $1$. 
We assume a set $\Pi$ of permissible path types is specified (by the RS designers) so that those path
types capture sensible connections between entities  \cite{POLLETIWICRS20}.

Suppose an RS suggests item $e_i$ to user $e_u$ (note that items and users are represented by entities in
the assumed auxiliary KG). A reason {\em for} this recommendation is simply taken to be a path $\pi \in \Pi$ 
that takes $e_{i}$ to $e_{u}$ in the KG. Thus we have an function $f$ that starts with the KG and 
the path $\pi$, takes inputs $e_i$ and $e_u$, and returns a set of reasons for the recommendation of 
$e_i$ to $e_u$. While this function can be implemented in several ways, in our implementation (described later) we employed
depth-first search in the KG \cite{POLLETIWICRS20}.

To illustrate, Figure \ref{fig:ex_red} shows through graphs an example where the recommendation of the  Red Phone  to a user is explained by the path $\pi_3 = (\mathrm{bought}, \mathrm{has}, \mathrm{has}^-)$, which goes through entities User, Laptop, Cutting Edge OS and Red Phone.

\input{images/example_phones}

%% file: images/example_phones.tex
\begin{figure}[t]
\hspace*{9mm}
\begin{subfigure}[!ht]{0.47\textwidth}
\begin{tikzpicture}[->,>=stealth',shorten >=1pt,auto,node distance=2.6cm, semithick,scale=0.9]
\tikzstyle{type0}=[align=center, rectangle, draw, rounded corners,
                     thin,bottom color=red, top color=red,
                     text=white, minimum width=2.5cm]
\tikzstyle{type1}=[rectangle, draw, rounded corners,
                     thin,bottom color=white, top color=white,
                     text=black, minimum width=2.5cm]
\tikzstyle{type2}=[align=center, rectangle, draw, rounded corners,
                     thin,bottom color=blue, top color=blue,
                     text=white, minimum width=2.5cm]
\tikzstyle{type3}=[align=center, rectangle, draw, rounded corners,
                     thin,bottom color=gray, top color=gray,
                     text=white, minimum width=2.5cm]
                     
\node[type0]         (A)        {\textbf{Red Phone}};
\node[type1]         (B)     at (5,0) {\textbf{Cutting Edge OS}};
\node[type3]         (C)     at (5, -2) {\textbf{Laptop}};
\node[type2]         (D)     at (0,-2) {\textbf{User}};

\path (A) edge             node {has} (B);
\path
(A) edge  [bend right, dashed] node {is recommended} (D)
(D) edge  [above, sloped]node {bought} (C)
(C) edge  [bend right]                 node {has} (B)
;
\end{tikzpicture}
\caption{Reason for recommending Red Phone.}
\label{fig:ex_red}
\end{subfigure}
 
\hspace*{9mm}
\begin{subfigure}[!ht]{0.47\textwidth}
\begin{tikzpicture}[->,>=stealth',shorten >=1pt,auto,node distance=2.6cm, semithick,scale=0.9]
\tikzstyle{type0}=[align=center, rectangle, draw, rounded corners,
                     thin,bottom color=green, top color=green,
                     text=black, minimum width=2.5cm]
\tikzstyle{type1}=[rectangle, draw, rounded corners,
                     thin,bottom color=white, top color=white,
                     text=black, minimum width=2.5cm]
\tikzstyle{type2}=[align=center, rectangle, draw, rounded corners,
                     thin,bottom color=blue, top color=blue,
                     text=white, minimum width=2.5cm]
\tikzstyle{type3}=[align=center, rectangle, draw, rounded corners,
                     thin,bottom color=gray, top color=gray,
                     text=white, minimum width=2.5cm]

\node[type0]         (A)        {\textbf{Green Phone}};
\node[type1]         (B)     at (4.8,0) {\textbf{Long Duration Battery}};
\node[type3]         (C)     at (5, -2) {\textbf{Laptop}};
\node[type2]         (D)     at (0,-2) {\textbf{User}};

\path (A) edge             node {has} (B);
\path
(A) edge  [bend right, dashed] node {is recommended} (D)
(D) edge  [above, sloped]node {bought} (C)
(C) edge  [bend right]                 node {has} (B)
;
\end{tikzpicture}
\caption{Reason for recommending Green Phone.}
\label{fig:ex_green}
\end{subfigure}
    \vspace*{-2ex}
\caption{Examples of reasons for and reasons against in item-based recommendation.}
\label{fig:ex_phones}
\end{figure}
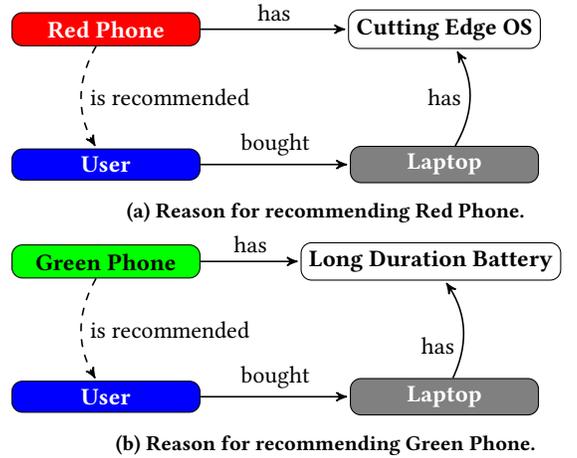

%% file: sections/32reasonsagainst.tex
\subsection{Reasons Against: What They Are, and How to Generate Them}
\label{sec:reasons_against}

We now focus on the main technical challenge in this work: how to generate
reasons {\em against} a particular recommendation. To do so, we resort
to the literature on practical reasoning in Philosophy, where we find Snedegar's 
rather comprehensive theory of reasoning \cite{snedegar2018reasons}. 
Snedegar presents five schemes by which reasons {\em against} can be generated
by an agent contemplating competitive options: 
\begin{description}
    \item[Scheme 1 (S1)]: a reason against an item $A$ is a reason for a competing option;
    \item[Scheme 2 (S2)]: a reason against an item $A$ is only a reason for NOT $A$ (not for any particular other option);
    \item[Scheme 3 (S3)]: a reason against an item $A$ is just a reason for the disjunction of the other options (say $B \lor C \lor D$);
    \item[Scheme 4 (S4)]: a reason against an item $A$ is a reason for each, i.e. all, of the alternatives to it.
    \item[Scheme 5 (S5)]: a reason against an item $A$ explains (or is part of the explanation as to) why $A$ promotes or respects some objective less well than 
    some other option.\footnote{This scheme requires one to specify a quantitative objective.}
\end{description}

These schemes have been defined by Snedegar at a highly abstract level; we must
taken them to a concrete level. We present our implementations in the
remainder of this section.

Our implementation of S1 generates a reason against a given item by generating reasons for other options. For instance, take the case where the RS has recommended two phones --- Red and Green --- as in Figure \ref{fig:ex_phones}. A reason against the Red Phone then would be that the Green Phone has a ``Long Duration Battery''. 

Scheme S2 is more delicate: how to define the negation of an item in the context 
of recommendations? The vague nature of this question led us to skip this scheme.

Our implementation of S3 goes through all competing options, collecting reasons for them that are not reasons for the option of interest; we then trim the list of reasons against to an arbitrary small number of reasons (e.g. 3). In our running example  we can imagine there is a Blue Phone and as reasons against the Red Phone we have that the Green Phone, the Blue Phone or both of them have long duration batteries. In practice S1 and S3 produce identical reasons against.

The implementation of S4 is similar to that of S3 to the extent that S4 takes reasons for all competing options into account (reasons against according to S4 are also reasons against according to S3). An example of reason against the Red Phone using S4 would be that both the Blue Phone and the Green Phone from the example above have adequate battery duration. 
The stringent nature of this scheme, where the intersection of reasons
is required, makes it hard to generate reasons against in practical
circumstances. 

Scheme S5 depends on a quantitative objective that can be the basis of explanations; 
this objective is used to determine whether a reason is for or against an option.
Consider in our phone example that the user has the objective of long battery life for her phone; with that piece of information, the RS can present the user with the reason against buying the Red Phone because it has a short duration battery.
We have implemented S5 by assuming that an objective function is known; however,
this is not a realistic assumption and future work should address the elicitation
of objectives at running time.

To illustrate the implemented algorithm, suppose an RS recommended $N$ items in an ordered set $\mathcal{I}: \{ i_{1}, i_{2}, ... i_{N} \}$ to user $u$. In Schema S1 (and S3) we define as reason  against an item $i_{r}$ the union of reasons for each of its alternatives $\mathcal{I} \backslash \{ i_{r} \}$ that are not reasons for $i_{r}$ itself. Hence we must iterate over the alternatives, extracting reasons for each one of them $\Phi \gets \Phi \cup \Phi_{u,i} \, \forall i \in \mathcal{I} \backslash \{ i_{r} \}$. Note that at this point we assume that function $f$, as described in Section \ref{sec:reasons_for}, is available. We then remove from $\Phi$ the reasons for our recommendation of interest, if any. The remaining reasons $\Omega = \Phi / \Phi_{u,i_{r}}$  are the reasons against $i_{r}$ -- as presented in the Algorithm \ref{algo:explanatory}.

\input{sections/algorithm}

Regarding the implementation of Schema 4 (S4), we follow a very similar procedure, except that instead of considering the union of reasons for its alternatives, we take the intersection. That is, we just replace the line $15$ of the Algorithm \ref{algo:explanatory} so as to take the intersection
of sets $\Phi \gets \Phi \cap \Phi_{u,i} \, \forall i \in \mathcal{I} \backslash \{ i_{r} \}$.

To close this section, consider an extended example using Scheme S1. 
We focus on Scheme 1 due to the fact that it captures most of the content
of Scheme S3 as well; as noted already, Scheme 2 does not seem conducive to a 
concrete implementation,
and Scheme 5 requires elicitation of user objectives --- finally, as discussed later
in connection with our experiments, Scheme 4 does not seem very promising in practice.

\begin{example}\label{example:USPedia}
We have built an RS to suggest University classes called Ganimedes.
A student asks for courses by presenting a few topics to Ganimedes; 
the RS then uses information from syllabuses and an associated knowledge
graph to produce recommendations. The knowledge graph, called  USPedia, 
collects information
about topics and their relationships; it was automatically harvested from
Wikipedia pages~\cite{POLLETIWICRS20}. 
We have defined a number of permissible paths for explanations 
(Section \ref{sec:reasons_for}).
For instance, one of them is 
$(\mathrm{subject}, \mathrm{broader}^-, \mathrm{broader})$;
as this permissible path indicates that a subject is of the same broader category as another topic of interest. That is, $\mathrm{subject}(X,Y)$
$\mathrm{broader}(Z,Y)$
$\mathrm{broader}(Z,W)$ means that Y is a topic of X, Z has the same broader categories of Y and W and, finally, that W is of the same broader category of a topic of X.

We assume that a course is likely to be about a given subject when it deals with topics that are related to that subject. 
For instance, a student who is a machine learning (ML) enthusiast 
would be satisfied with a course that is about statistical models 
even if the course is not focused on ML itself.

\input{images/example_responses}

Figure \ref{fig:explanations} conveys a number of explanations 
generated by our RS. In this case, the student asked our RS for courses about \textit{Stochastic Resonance}, and was suggested classes with codes PME3430 and PME3479. The RS found two reasons for PME3430 (Fig. \ref{fig:ex1} and Fig. \ref{fig:ex2}) and one for PME3479 (Fig. \ref{fig:ex3}). Note that both recommendations share the \textit{reason for} depicted in Fig. \ref{fig:ex2} and \ref{fig:ex3}, thus it cannot be a reason against for none of them. On the other hand, the one in Fig. \ref{fig:ex1} is a reason for only PME3430; therefore, it is a reason against PME3479.
\hfill $\Box$
\end{example}

%% file: sections/algorithm.tex
\begin{algorithm}[!ht]
    \caption{Explanation Generation using Scheme S1}\label{euclid}
    \label{algo:explanatory}
    \begin{algorithmic}[1]
        \Procedure{reasons-for}{$i, u, \Pi, \mathcal{G}$}
            \State $\Phi_{u,i}= \{\}$ \Comment{Set of reasons for $i$}
            \For{\textbf{all} $\pi \in \Pi$}
                \State $\phi$ $\gets$ $f(u,i, \pi | \mathcal{G} )$ \Comment{Function described in Section \ref{sec:reasons_for}}
                \State $\Phi_{u,i}$ $\gets$ $\Phi_{u,i} \cup \phi$
            \EndFor
            \State \Return $\Phi_{u,i}$
        \EndProcedure
        \Procedure{reasons-against-S1}{$i_{r}$, $u$, $\mathcal{I}, \Pi, \mathcal{G}$}
            \State $\Omega_{u,i_{r}}$ $\gets$ $\{\}$ \Comment{Set of reasons against $i_{r}$}
            \State $\Phi = \{\}$
            \State $\Phi_{u,i_{r}}$ $\gets$ \textsc{reasons-for}($i_{r}$, $u$, $\Pi$, $\mathcal{G}$) \Comment{Set of reasons for $i_{r}$}
            \For{$i \in \mathcal{I} \backslash \{i_{r}\}$} \Comment{Iterate over $i_{r}$ alternatives}
                \State $\Phi_{u,i}$ $\gets$ \textsc{reasons-for}($i$, $u$, $\Pi$, $\mathcal{G}$)
                \State $\Phi$ $\gets$ $\Phi \cup \Phi_{u,i}$
            \EndFor
        \State $\Omega_{u,i_{r}}$ $\gets$ $\Phi \backslash \Phi_{u,i_{r}}$
        \State \Return $\Omega_{u,i_{r}}$
        \EndProcedure
    \end{algorithmic}
\end{algorithm}

%% file: images/example_responses.tex
\newcommand\mycolor{black}
\begin{figure}[!ht]
\centering
\begin{subfigure}[!ht]{0.95\linewidth}
\begin{tikzpicture}[->,>=stealth',semithick,scale=0.85]
\tikzstyle{type0}=[align=center, rectangle, draw, rounded corners,
                    thin,bottom color=\mycolor, top color=\mycolor,
                    text=white, minimum width=1cm]
\node[type0]         (A)    {\textbf{PME3430}};
\node[type0]         (B)     at (0,2) {\textbf{Robotic Sensing}};
\node[type0]         (C)     at (5, 0) {\textbf{Sensorial System}};
\node[type0]         (D)     at (5,2) {\textbf{Stochastic Resonance}};
\path (A) edge [right] node {subject} (B);
\path
(C) edge [right] node {broader} (B)
(C) edge [right] node {broader} (D);
\end{tikzpicture}
\caption{Reason for PME3430 and against PME3479.}
\label{fig:ex1}
\end{subfigure}

\begin{subfigure}[!ht]{0.95\linewidth}
\begin{tikzpicture}[->,>=stealth',shorten >=1pt,auto,node distance=2.6cm, semithick,scale=0.85]
\tikzstyle{type0}=[align=center, rectangle, draw, rounded corners,
                     thin,bottom color=\mycolor, top color=\mycolor,
                     text=white, minimum width=1cm]

\node[type0]         (A)        {\textbf{PME3430}};
\node[type0]         (B)     at (0,2) {\textbf{Auditive System}};
\node[type0]         (C)     at (5, 0) {\textbf{Sensorial System}};
\node[type0]         (D)     at (5,2) {\textbf{Stochastic Resonance}};

\path (A) edge [right] node {subject} (B);
\path
(C) edge [right] node {broader} (B)
(C) edge [right] node {broader} (D) ;
\end{tikzpicture}
\caption{Reason for PME3430.}
\label{fig:ex2}
\end{subfigure}

\begin{subfigure}[!ht]{0.95\linewidth}
\begin{tikzpicture}[->,>=stealth',shorten >=1pt,auto,node distance=2.6cm, semithick,scale=0.85]
\tikzstyle{type0}=[align=center, rectangle, draw, rounded corners,
                     thin,bottom color=\mycolor, top color=\mycolor,
                     text=white, minimum width=1cm]

\node[type0]         (A)        {\textbf{PME3479}};
\node[type0]         (B)     at (0,2) {\textbf{Auditive System}};
\node[type0]         (C)     at (5, 0) {\textbf{Sensorial System}};
\node[type0]         (D)     at (5,2) {\textbf{Stochastic Resonance}};

\path (A) edge [right] node {subject} (B);
\path
(C) edge [right] node {broader} (B)
(C) edge [right] node {broader} (D) ;
\end{tikzpicture}
\caption{Reason for PME3479.}
\label{fig:ex3}
\end{subfigure}
\caption{Examples of S1 with two reasons for and against the recommendation PME3430.}
\label{fig:explanations}
\end{figure}
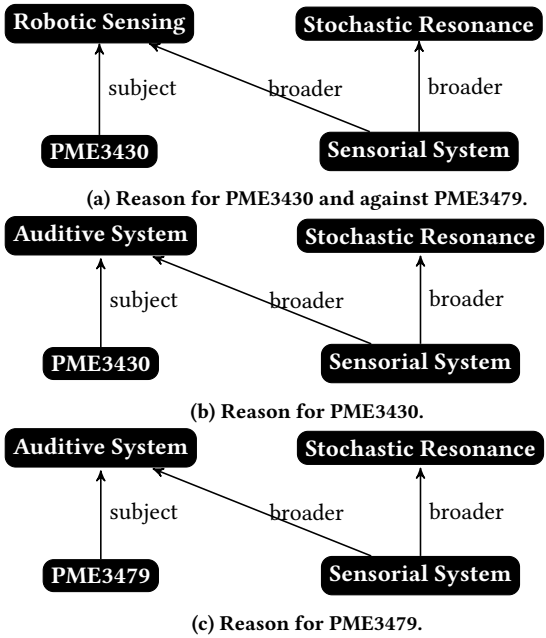


%% file: sections/4experiments.tex
\section{Experiments}

In this section we describe experiments with simulated
and real users; we first examine the feasibility of
our techniques in Section \ref{subsection:Simulated}
and then we discuss the reaction of human users to our
approach in Section \ref{subsection:Human}.

\subsection{Evaluation of feasibility: simulated interactions}\label{subsection:Simulated}

We have first evaluated our proposal from two perspectives: (1) the fraction of recommendations for which we can find at least one explanation (we refer to it as {\em coverage}) and (2) the average number of reasons we can find to support/attack a given recommendation (we refer to it as {\em support})~\cite{POLLETIWICRS20, Zhang2019}. These metrics offer a glimpse at the workings of our proposal in a real-world scenario from a objective perspective.
To carry out our experiments, we trained an RS based on TransE~\cite{Bordes2013} embedding from the USPedia knowledge graph employed in Example 3.1, using the same set-up as in Ref.~\cite{POLLETIWICRS20}. We built our simulated interactions by asking for the Top-$4$ recommendations of randomly sampled $100$ cases. Next, for each interaction, we used our proposed method to retrieve both reasons for and against.

Regarding \textit{reasons for}, Table  \ref{tab:objective_analysis} shows that we obtained $79.33\%$ coverage and a support mean of $2.0$, similar results to those reported in previous works~\cite{POLLETIWICRS20,Zhang2019}. As for \textit{reasons against}, we ran our experiments considering Schemas S1 and S4. Both the coverage ($85.1\%$) and support ($2.3$) obtained for S1 are higher than those from \textit{reasons for}. This result was expected since S1 implementation considers more aggregated reasons for alternatives than it removes from the recommendation being explained. 

On the other hand, Scheme S4 could {\em not} generate a single reason against at all (coverage $0\%$!). As Scheme S4 requires that a reason against an option must be a reason for all of its alternatives, it imposes a restriction so rigorous that it is in fact unfeasible in practice.

\begin{table}[ht]
    \centering
    \begin{tabular}{l|r|c}
    \toprule
    \textbf{Explanation Type} & \textbf{Coverage} & \textbf{Support} \\
    \midrule
    \textbf{Reason For}            & $79.3\%$ & $2.0 \pm 1.0$ \\
    \textbf{Reason Against (S1)}   & $85.1\%$ & $2.3 \pm 1.4 $\\
    \textbf{Reason Against (S4)}   & $0\%$ & -  \\
    \bottomrule
    \end{tabular}
    \caption{Coverage and Support for reasons for and reasons against using Schemas S1 and S4.}
    \label{tab:objective_analysis}
\end{table}

\subsection{Evaluation with Human Subjects}
\label{subsection:Human}
One could expect the fact that an RS can be built with reasons for/against does
not mean that human subjects would be satisfied with it; to determine whether
indeed our approach is a valuable one, we carried out an experiment to address
the following questions: \\
1) Do reasons for/against have value for users? \\
2) Do reasons against reduce an RS persuasion? \\
3) Do users perceive a conflict of interest in their interaction with an RS? \\
4) Do reasons for/against influence user choices? \\

\begin{figure*} 
    \includegraphics[width=0.9\textwidth]{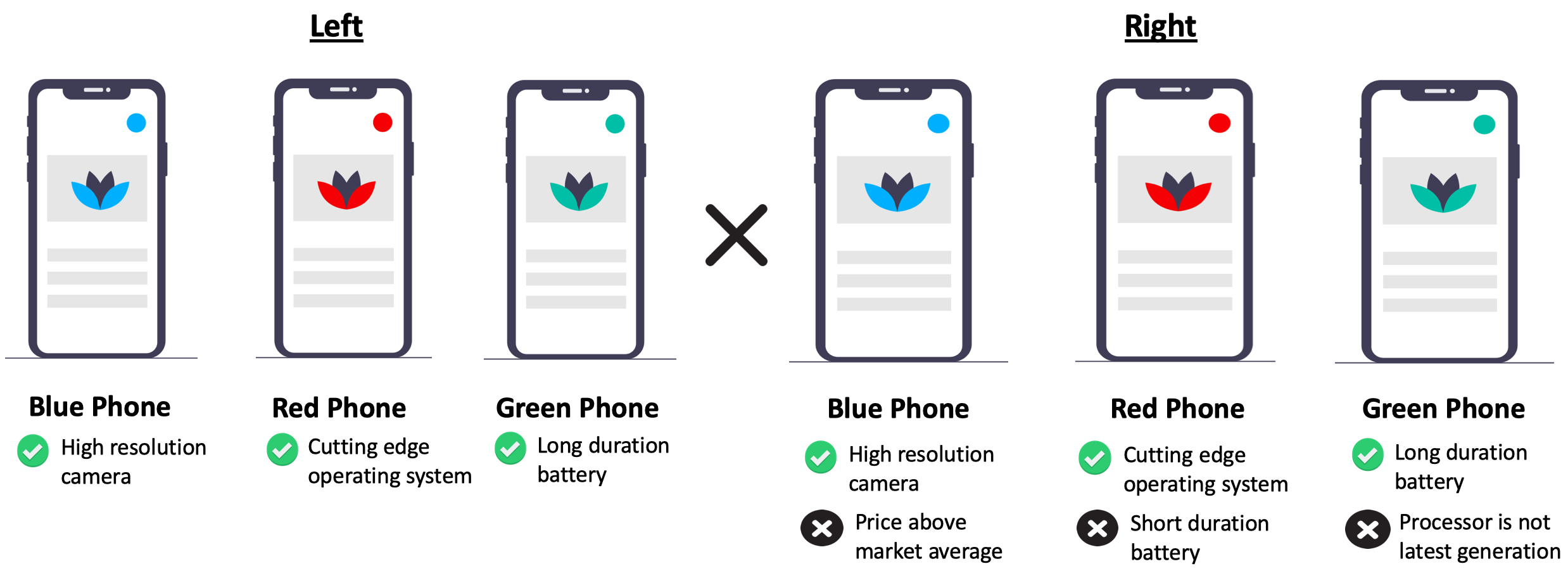}
    \caption{Experiment: just one reason for (left); one reason for and one reason against (right).}
    \label{fig:exp_comparison}
\end{figure*}

Our experiment took $31$ subjects, all of which are undergraduate students,
and asked them to evaluate two RS implementations, one displaying only reasons for recommendations,
and the other displaying reasons for and against them.
Subjects were presented with an e-commerce mock-up where they received recommendations
concerning smartphones.
Each subject first received a recommendation and one reason for, and
was asked to select an item; 
then the subject received a recommendation with one reason for and one reason against,
and was again asked to select an item.
Note that we avoided
presenting too many reasons at once.
Figure \ref{fig:exp_comparison} depicts the information presented.

Each subject then evaluated the two RSs individually using five 
{\em explanation metrics}~\cite{Tintarev:2007:SER:1547550.1547664} 
that are presented in Table~\ref{tab:quest}. Each subject ranked each
RS with respect to each explanation metric using a survey-based Likert 
psychometric scale~\cite{likert1932} from 1 to 5 (standing for 
``Strongly disagree'', 2 ``Disagree'', 3 ``Neither agree nor disagree'', 
4 ``Agree'', and 5 ``Strongly agree'').
This scale was used to reduce central tendency and social desirability
biases where subjects do not want to be identified with extreme positions.
Finally, each subject could write a short free text with thoughts about
the RSs. 

\begin{table}
    \centering
    \begin{tabularx}{\linewidth}{l|X}
    \toprule
    \textbf{Metric} & \textbf{Question} \\
    \midrule
        transparency & The explanation on the right helped me understand why the items were recommended better than the explanation on the left \\
        persuasion & Based on the explanation on the right, I was more prone to follow the recommendation than based on the explanation on the left \\
        engagement & The explanation on the right helped me learn more about the recommended items than the explanation on the left \\
        trust & The explanation on the right contributed more to increase my confidence in the recommendations than the explanation on the left \\
        effectiveness & The explanation on the right made me more confidence about making the best choice than the explanation on the left \\
    \bottomrule
    \end{tabularx}
    \caption{The five explanation metrics that subjects had to take into account in the experiment.}
    \label{tab:quest}
\end{table}

\begin{figure}[!ht]
     \centering
    \begin{subfigure}[!ht]{0.5\textwidth}
    \centering
    \includegraphics[width=\textwidth]{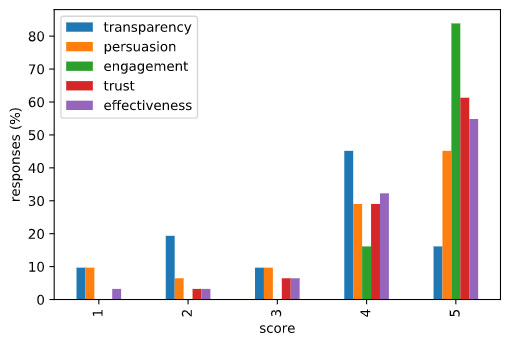}
    \caption{Visual representation for explanation metrics average scores.}
    \label{fig:visual}
    \end{subfigure}
     \hfill
    \begin{subfigure}[!ht]{0.5\textwidth}
         \centering
         \includegraphics[width=\textwidth]{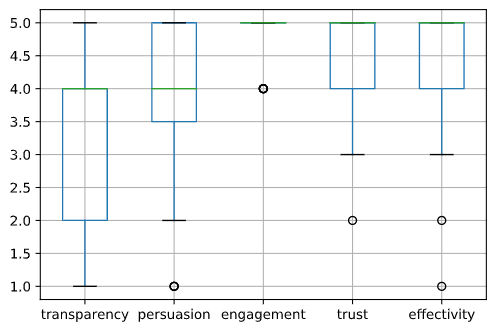}
        \caption{Boxplots for the explanation metrics.}
        \label{fig:boxplot}
     \end{subfigure}
     \caption{Results from the experiment with human subjects.}
     \label{fig:2f}
\end{figure}

Figure \ref{fig:visual} shows the percentage of responses given by subjects. Responses, notably for \textit{engagement}, \textit{trust} and \textit{effectiveness}, are concentrated around scores $4$ and $5$. This result indicates that users mostly agree that showing reasons against a recommendation adds value  with respect to trust, engagement and effectiveness of RS.
Figures \ref{fig:visual} and \ref{fig:boxplot}  show that there was a divergence amongst users about whether the proposed explanation paradigm increases transparency. As our method is model-agnostic (it makes no assumptions about the RS internal behavior), the explanations were unable to shed light on how items were actually recommended. As the transparency score peaked around $3$, this does not mean reasons for/against were adverse to transparency; it means that they were  as good as just reasons for.

We expected a possible drawback of our proposal would be a reduction in persuasion (as reasons against might make the users less likely to follow recommendations). 
Figure \ref{fig:boxplot} shows that the down whisker is longer for persuasion than it is for trust, engagement and effectiveness. However, note that the boxplot for persuasion is skewed up; thus most users felt more convinced when reasons against were present. By doing a further analysis of textual comments, we found out that persuasion increases are produced by higher trust in the RS. Consider two comments:

\begin{displayquote}
1) \textit{I always think that recommendations that bring positive and negative aspects are fairer, and could influence me more into buying the product, once I feel I am not being misled.} \\
2) \textit{As the first example [the first RS] shows only strong points for each product, it leads the user to have a certain mistrust about the suggestions.}
\end{displayquote}

Comments also indicate that many users expect the RSs to try to lead them into a decision,
sensing a conflict of interest in the process. Consider the following comment:

\begin{displayquote}
3) \textit{Differently from marketing which always idealize the product, this one seems to show the reality about it, thus I feel I understand the recommended product in its real form.} 
\end{displayquote}

These comments corroborate our hypothesis that, indeed, reasons against have a significant positive impact on the user decision-making process. As a matter of fact, a full $45\%$ of our test subjects changed their choices after we presented reasons against.

%% file: sections/5conclusion.tex
\section{Conclusion}\label{section:Conclusion}

In this paper we have proposed a novel feature for RSs, whose
goal is to enhance trust by acting responsibly; namely, we
investigated the generation of reasons {\em for} and {\em against}
recommendations. By displaying such reasons, an RS not only
helps the user to reach the most rewarding decision, but the
RS acts on its own interest in building trust. 

We have developed ways to generate reasons for/against using
an auxiliary KG by adapting Snedegar's theory of practical 
reasoning. Our implementation demonstrates that additional
calculations needed to generate such reasons do not affect overall
performance. By implementing Snedegar's theory we have
found difficulties with some of his schemes for reasons against;
we suggest that his Scheme 1 is the most appropriate in practice
at the moment. Moreover, our experiment with human subjects  
demonstrated that reasons against can significantly increase trust,
engagement, and even persuasion. Overall we demonstrated that adding
reasons against items does improve RSs. 

Future work should investigate how much information should
be given to users when presenting reasons for/against.
It would also be useful to explore mental models
of the user so as to extract quantitative objectives to use
in Snedegar's Scheme S5. Moreover, it would be important
to evaluate our proposals at scale.

%% file: sections/6acknowledgements.tex
\section{Acknowledgments}
This work was carried out with the support of Ita\'u Unibanco S.A.; the second author has been supported by the Ita\'u Scholarship Program (PBI), linked to the Data Science Center (C2D) of the Escola Polit\'ecnica da Universidade de S\~ao Paulo. The third author has been partially supported by the Conselho Nacional de Desenvolvimento Cient\'ifico e Tecnol\'ogico (CNPq), grant 312180/2018-7. This work has been partially supported by the Funda\c{c}\~ao de Amparo \`a Pesquisa do Estado de 
S\~ao Paulo (FAPESP), grant 2019/07665-4. We acknowledge support by CAPES finance code 001.
We are also grateful to the Center for Inovation at Universidade de S\~ao Paulo (InovaUSP) for hosting our lab.